\pdfoutput=1

\documentclass[11pt]{article}

\usepackage{EMNLP2023}

\usepackage{times}
\usepackage{latexsym}
\usepackage{graphicx}
\usepackage{amsmath,amsfonts,amssymb}
\graphicspath{ {emnlp2023-latex/} }
\usepackage{array}
\newcolumntype{H}{>{\setbox0=\hbox\bgroup}c<{\egroup}@{}}

\newcommand{\metricnospace}[0]{Conditional Informativeness}
\newcommand{\metric}[0]{\metricnospace\ }

\usepackage[T1]{fontenc}

\usepackage[utf8]{inputenc}
\usepackage{caption}
\usepackage{subcaption}
\usepackage{tabularx}

\usepackage{microtype}

\usepackage{inconsolata}
\usepackage{multirow}

\title{Large Language Models as Annotators: 
Enhancing Generalization of NLP Models at Minimal Cost}

\author{Parikshit Bansal \\
  Microsoft Research India \\
  \texttt{parikshitb52@gmail.com} \\\And
  Amit Sharma \\
  Microsoft Research India \\
  \texttt{amshar@microsoft.com} \\}

\begin{document}
\maketitle

\begin{abstract}
State-of-the-art supervised NLP models achieve high accuracy but are also susceptible to failures on inputs from low-data regimes, such as domains that are not represented in training data. As an approximation to collecting ground-truth labels for the specific domain, we study the use of large language models (LLMs) for annotating inputs and improving the generalization of NLP models. 
Specifically, given a budget for LLM annotations, we present an algorithm for sampling the most \textit{informative} inputs to annotate and retrain the NLP model. 
We find that popular active learning strategies such as uncertainty-based sampling do not work well.  Instead, we propose a sampling strategy based on the difference in prediction scores between the base model and the finetuned NLP model, 
utilizing the fact that most NLP models are finetuned from a base model. Experiments with classification (semantic similarity) and ranking (semantic search) tasks show that our sampling strategy leads to significant gains in accuracy for both the training and  target domains.
\end{abstract}

\section{Introduction}
\label{sec:intro}
A common limitation of supervised NLP models is that they fail to generalize in \textit{low data} regimes, corresponding to inputs from subgroups or domains that have limited labelled data in the training set. 
These generalisation errors occur due to distribution shifts between the new inputs and training data, that render some of the correlations learnt by the model as invalid~\cite{wang2022generalizing}. For instance, models may learn spurious correlations with sensitive attributes like gender~\cite{sun2019mitigating} or may over-emphasize lexical patterns~\cite{gururangan2018annotation}; or in some cases, inputs may exhibit a new concept that has not been seen in the training data~\cite{10.1145/2523813}. 

As a motivating example, consider the task of determining \textit{semantic similarity} between a pair of sentences~\cite{reimers2019sentence}. This task forms the basis of information retrieval and recommendation systems such as similar question recommendation on online forums~\cite{wang-etal-2018-glue} or product recommendation on e-commerce websites~\cite{he2016ups}. In such systems, 
it is common to encounter new unseen domains during deployment. 
For instance, introduction of a new item category  or  users from a new demographic may cause failures for a deployed model due to a shift in distribution of inputs in the system compared to the training data. Unlabelled data is readily available for such distribution shift (i.e., new questions posted by users or  items from a new category), but labelling the data requires considerable human effort. In other cases, failures may occur due to hard-to-learn semantic patterns found in a small minority of the training data (see the example pair containing lexically similar questions on oxygen and glucose in Figure 1).


A common solution in all these cases is to collect more labelled data distinct from the distribution of training data, but labelling (or \textit{annotating}) data is an expensive and manual process. To address this issue, prior work suggests using large language models (LLMs,~\cite{ouyang2022training,brown2020language}) to annotate data. LLMs like GPT-3 obtain promising accuracy for annotating data for a variety of NLP tasks including sentiment classification~\cite{ding2022gpt}, keyword relevance~\cite{choi2023chatgpt,gilardi2023chatgpt} and question answering~\cite{gilardi2023chatgpt}. However,  LLM-based annotations can be noisy and due to efficiency reasons, we cannot deploy LLM models directly.

In this paper, we take the natural next step and ask whether annotations from LLMs can be used to enhance generalization of existing NLP models. 
Given a corpus of unlabelled data, we find that a naive application of LLMs (annotating inputs at random) provides only marginal gains on total accuracy and in some cases, can worsen accuracy for low-data groups. To optimize the sampling, we formulate the problem of sampling inputs for annotation as an \textit{active learning} problem~\cite{zhang2022survey}. However, we find that the popular sampling strategy based on model uncertainty~\cite{lewis1995sequential} is also not optimal for LLM-based annotation. 

Using experiments on  classification (semantic similarity) and ranking (semantic search) tasks, we propose an alternative strategy for sampling inputs. 
For cost-efficient sampling of new unlabeled inputs for LLM annotations, an intuitive solution is to annotate only those inputs that 
the NLP model is expected to be incorrect on, i.e., inputs where the NLP model's prediction and the ground truth label would differ.
In the absence of GT labels for new inputs, we propose a metric, \textit{\metricnospace}, that approximates this intuition. We utilize the fact that state-of-the-art supervised NLP models are often finetuned from a \textit{base model} such as BERT~\cite{vaswani2017attention} that provides an initial embedding for the input.  
For a given input and an NLP task, \metric measures the deviation between the prediction score from the base model and the score from the NLP  model finetuned using the available labelled data for the task. We argue that the inputs which have the max deviation between the two scores are the ones likely to be incorrectly predicted by the finetuned model and hence the most informative ones for finetuning over the base model. 

Our sampling metric provides a practical way to improve generalization of NLP models for a task (see Figure 1 for an illustration). Given a budget for LLM annotation (i.e., number of queries), we select the inputs having the maximum \metric for LLM annotation and then retrain the NLP model using this additional training data. Our algorithm shows significant improvements in target domain and total accuracy, on the Quora dataset for the semantic similarity task, and on Amazon and Wikipedia datasets for the semantic search task. Our algorithm also provides higher gains than the uncertainty-based sampling from active learning literature. This may be because of the error distribution of LLM annotations: only for inputs with high deviation, the LLM-based  annotations may be expected to be more accurate than the base model. 



To summarize, we make the following contributions. \\
1) The \metric metric for sampling inputs for LLM-based annotation that outperforms commonly used active learning approaches.\\
2) Experiments on semantic similarity and search tasks that show LLM annotations can significantly improve both in-domain and target domain accuracy.

\begin{figure*}[ht]
\centering 
\includegraphics[width=\textwidth]{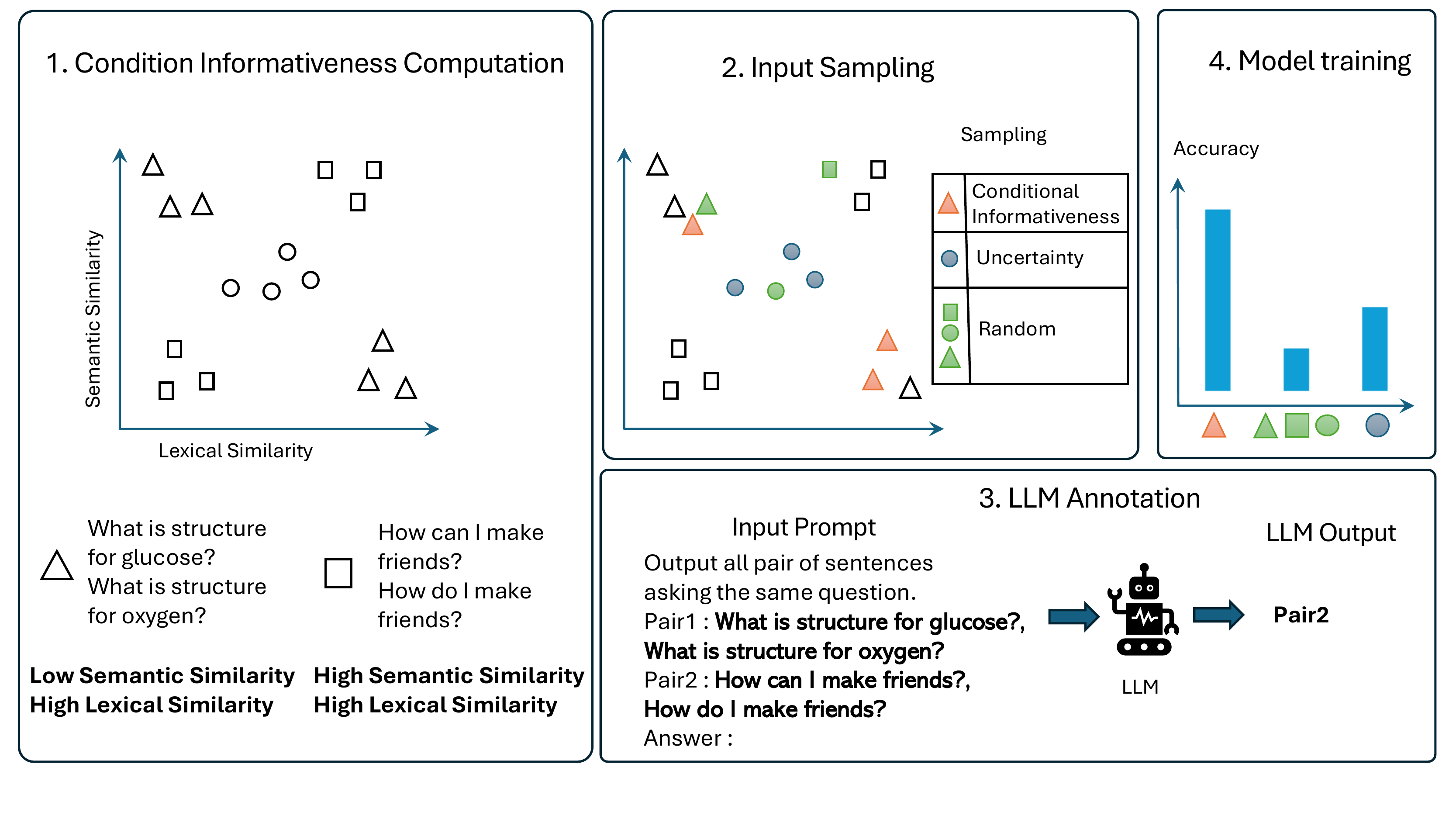}
\vspace*{-30pt}
\caption{Enhanced Generalization using LLM Annotations. Illustration of our algorithm using the duplicate question detection task. We propose a  sampling strategy based on deviation of an NLP model's similarity score from the base model, called  \textit{(base model)}-conditional informativeness. Inputs are sampled using this strategy (Step 2), annotated using an LLM (Step 3) and then added to the training set of the NLP model. Our sampling strategy performs significantly better than random or active learning-based strategies.}
\label{fig:fig1}
\end{figure*}

\section{Related Work}
\label{sec:related_work}
\paragraph{LLMs for data augmentation.} 

A popular framework for improving a NLP model's generalization  has been to generate new data using LLMs and test the model's output using a human-in-the-loop, i.e. LLMs are used in partnership with human participants for data generation and testing/debugging models ~\cite{ribeiro-lundberg-2022-adaptive,wang2021want}. In recent work, \cite{he2023targeted} utilise the same strategy for training an NLP model: they use GPT-3 for data generation over under-represented groups, which are then annotated by users before including in training set. 

However, with more capable LLMs like ChatGPT, LLMs are now capable of not just generating data, but also annotating it (while faithfully following annotation instructions). 
Recent work ~\cite{gilardi2023chatgpt,he2023annollm,ding2022gpt} has looked at the annotation accuracy for LLMs and found them to be at par with crowd-worker annotators. Combining generation and annotation, parallel to us, \cite{whitehouse2023llm} explore the utility of both input and labels generated from  LLMs for crosslingual common sense reasoning tasks. Similarly, for the task of building a sentence embedding using contrastive learning, \cite{cheng2023improving} use LLMs to both generate novel input pairs and then score their similarity.

Motivated by real-world applications from information retrieval, we focus our attention on the  \textit{unsupervised domain adaptation}\cite{ramponi2020neural} (UDA) setting where unlabelled inputs are easily available. UDA methods assume a source labeled domain and a target unlabeled domain with the goal of adapting to the target domain (while also performing well on the source domain). For instance, \cite{saad2023udapdr} motivate the passage reranking task where a large number of unlabelled passages are available. They use LLMs to  generate synthetic queries for a given passage and then use such augmented data to train a downstream model. 
Given the potential of LLM-annotated data for training downstream classifiers and the associated costs of querying them, we study how to  \textit{efficiently} utilise these annotations to train a more generalizable NLP model; specifically, which inputs to annotate for maximum benefit? 

\paragraph{Semantic similarity with limited labeled data.} 
~\cite{chen2023empirical} present a comprehensive survey of data augmentation techniques for limited label data settings in NLP. 
AugSBERT~\cite{thakur2020augmented} present an augmentation strategy that uses a bigger (oracle) cross-encoder model for generating (pseudo-)labels for unlabeled inputs. These inputs are then utilised to train a smaller and efficient NLP model. Such an oracle, however, is limited by the training data whereas LLMs are known to have zero-shot capabilities  that generalize to new domains\cite{hou2023large}.

In addition to augmentation, unsupervised domain adaptation methods have also been proposed. 
Apart from the main task learning loss, 
\cite{ramesh-kashyap-etal-2021-domain} propose  an additional loss which minimizes the divergence between source and target domain representations. Recent work UDApter~\cite{malik2023udapter} combines UDA methods with adapters for efficient domain adaptation.  However, domain matching techniques work only  under a restrictive set of assumptions~\cite{li2020rethinking}. Instead, we aim to approximate the ground-truth labels through LLMs, thereby converting the unsupervised problem into a simpler, supervised learning problem.
\cite{dua2022adapt} investigates the failure modes of Open-domain question answering when faced with distribution shifts. In addition they propose a few-shot data augmentation method for improving generalisation of these models. The augmentations uses LLMs to generate question for a given passage.

\paragraph{Active Learning.} 
Choosing which inputs to annotate has been classically studied as an active learning problem~\cite{settles2009active}.In active learning setup, we are given a small set of  $L$ labeled inputs, along with a large pool of $U$ unlabeled inputs. We are also specified a budget $B$, which denotes the number of inputs from the unlabeled data that can be annotated by an oracle/human. 
Active learning explores how to best sample $B$ inputs from the unlabeled pool to maximize the generalization accuracy of the final model that is trained on the original $L$ + (annotated) $B$ samples. 
{Active Learning} uses two primary criterion for sample selection : Informativeness and Representativeness \cite{zhang2022survey}. The most popular informativeness technique is uncertainty sampling ~\cite{lewis1995sequential,schroder2021revisiting} and for representativeness is diversity/density.
As an application, recent work \cite{margatina2023active} uses active learning in an in-context learning setting for LLMs and shows that similarity based sampling (instead of uncertainty and diversity) are most effective for in-context learning. 
In this paper,  we focus on LLM-based annotations and  evaluate  the uncertainty-based informativeness sampling technique. Based on our experiments, we also propose a new informativeness criterion.

\section{Conditional informativeness criterion for sampling LLM annotations}
\label{sec:motivation}
\subsection{Background: Building NLP classifiers using base models}
Given a domain of sentences, $\mathcal{X}$, and a task $\mathcal{T} : \mathcal{X} \to \{0,1\}$ we consider learning a classifier function $f : \mathcal{X} \to \{0,1\}$ which follows the task i.e. $f(x) = \mathcal{T}(x) \ \forall\ x \in \mathcal{X}$. 
The function aims to learn features which are predictive of the output label and their mapping to the output label. 
A subset of the domain $\mathcal{X}$ is denoted by $X=\{x_0,x_1,x_2,\ldots,x_{|X|}\} \subseteq \mathcal{X}$. The output label of $x_i$ is $\mathcal{T}(x_i)$ and is denoted by $t_i$. A set of examples can be represented as 
\begin{equation}
    D = \{(x_i,t_i) : i \in [|X|]\}
\end{equation}
Unlabeled examples lack the task label $t_i$.

\paragraph{Semantic Similarity.} As an example, consider the semantic similarity task~\cite{cer2017semeval}. 
Inputs for semantic similarity come from $\mathcal{X} \times \mathcal{Y}$ where $\mathcal{X}$ and $\mathcal{Y}$ are a pair of domain of sentences. The domains can be the same or different for symmetric and asymmetric similarity respectively. For a given input $(x_i,y_i)$, the task output is 1 if a pair are semantically similar, and 0 if they are not. The classifier for semantic similarity is hence defined as $f : \mathcal{X}\times \mathcal{Y} \to \{0,1\}$. We denote a training set as  : 
\begin{equation}
    D = \{((x_i,y_i),t_{i}) : i \in [|X|]\}
\end{equation}
Further details on semantic similarity are in Supp.~\ref{app_sec:formanlising_ss}. 

\paragraph{Finetuning on Base model.} NLP models are usually finetuned on top of some pretrained text models (e.g., we use \texttt{MSMARCO-DistilBERT-v4} for semantic similarity) called \textit{Base} model. The base model adheres to an approximation of the task based on the pretraining dataset and provides initial embedding for the input. 
We call these features defined by the base model as \textit{pretrained features}. 




\subsection{A domain adaptation case study: Which inputs to annotate?}
To evaluate different input sampling techniques for LLM annotations,  
we consider the semantic similarity task of duplicate question detection. We train bi-encoders (SBERT~\cite{reimers2019sentence}) on the Quora Questions Pair dataset~\cite{wang-etal-2018-glue},using \texttt{MSMARCO-DistilBERT-v4} as the base model.
To simulate a challenging target domain, we remove 60\% of  ``extreme'' examples from the training dataset. These are examples where the base model either obtains the lowest mean squared error w.r.t. the ground truth labels or obtains the highest mean squared error.  That is, half of the examples (30\%) are the \textit{easy} examples the base model is (most) correct on and the remaining half are the \textit{hard} examples where the base model is (most) incorrect on. Further, we remove labels from the target domain.
Hence from the original data we have 40\% ``source'' labeled examples and 60\%  ``target'' unlabeled examples. For accuracy evaluation on both source and target domains, we create analogous domains over the test set too.


We consider an active learning setup where selected inputs from target domain can be annotated by an LLM and augmented in the training set. After augmentation, the model is trained on the source domain + augmented dataset. 
We consider two popular active-sampling approaches in literature: Random and Uncertainty-based sampling. Apart from these, we include two additional sampling techniques based on our knowledge of the target domain: \textit{base-consistent-sample} and \textit{base-inconsistent-sample}. These are designed to capture the \textit{easy} and \textit{hard} examples that constitute the target domain.
Given labeled data L, unlabeled inputs U and a budget for annotation as B we have : 
\begin{itemize}
    \item  \textbf{random-sampling.} We randomly select $B$ inputs out of the $U$ unlabeled inputs for annotations.
    \item \textbf{uncertainty-sampling. } We first finetune the base model on the given labeled data $L$ and then select the $B$ (budget) most uncertain (according to the finetuned model) unlabeled inputs (out of $U$). 
    \item \textbf{base-consistent-sampling.} We choose top $B$ examples having lowest (MSE) error on base model predictions with GT labels.
    \item \textbf{base-inconsistent-sampling}. We choose top $B$ examples having highest (MSE) error on base model predictions with GT labels.
\end{itemize}
These $B$ inputs are then annotated and included for final training on $L+B$.

\paragraph{AUC under different sampling strategies. } 
Using gpt-3.5-turbo as the annotater LLM, we report AUC (area-under-(ro)curve) in Table~\ref{tab:quora_motivation}. We set a budget $B$ of 10\% of the dataset for annotation.
For details on prompts used, see Sec~\ref{subsec:finalalgo}.

Looking at the AUC metric for the complete target domain, we observe that random-sampling and uncertain-sampling lead to similar improvements compared to the training set. Compared to these active learning techniques, base-inconsistent-sampling leads to almost twice the AUC improvement.   That is, annotations with LLM are best under base-inconsistent-sampling. Remarkably, with only 10\% of the examples annotated, AUC with base-inconsistent-sampling is even higher than the setting where we augment the \textit{full} target domain (100\% of examples).  In contrast, base-consistent-sampling hurts generalization. Even though  base-consistent-sampling was designed to sample examples with low base model error, it obtains worse AUC than  base-inconsistent-sampling on test examples with low base model error. 
Results on using Ground Truth (GT) labels for annotations (instead of LLM annotations)  are in Supp. Table~\ref{tab:quora_motivation_gt}. 


\begin{table*}[h]
\centering
\begin{tabular}{|l|rrr|}
\hline
Data & Complete Test & High Base Error & Low Base Error  \\
\hline

Initial Train Set & 86.824 $\pm$ 0.038 & 59.335 $\pm$ 0.139 & \textbf{99.068 $\pm$ 0.048}\\
\hline
+ 100\% (complete target domain) & 87.544 $\pm$ 0.035 & \textbf{65.785 $\pm$ 0.121} & 98.164 $\pm$ 0.044 \\
\hline
+ Random-sampling 10\% & 87.052 $\pm$ 0.151 & 60.551 $\pm$ 0.701 & 98.805 $\pm$ 0.058\\
+ Uncertain-sampling 10\% & 87.620 $\pm$ 0.029 & 61.594 $\pm$ 0.433 & 99.081 $\pm$ 0.024\\
+ Base-consistent-sampling 10\% & 86.763 $\pm$ 0.149 & 59.986 $\pm$ 0.340 & 98.833 $\pm$ 0.024\\
+ Base-inconsistent-sampling 10\% & \textbf{88.108 $\pm$ 0.062} & \textbf{65.538 $\pm$ 0.175} & 98.861 $\pm$ 0.046 \\
\hline
\end{tabular}
\caption{\label{tab:quora_motivation}
AUC for Quora duplicate questions task, before and after including LLM-based annotations using four different sampling techniques: random, uncertainty, base-consistent and base-inconsistent. AUC is evaluated on the full test set, the test subset with high base model error and the test subset with low base model error. 
Sampling just 10\% of the data for annotation using base-inconsistent-sampling is better than annotating with the complete  (100\%) target dataset.
}
\end{table*}

\paragraph{Implications.}
The above results indicate that for LLM annotations,  uncertainty-sampling may not be the best technique.
To understand these results, note that 
 the original model finetuned on training set (first row in Table 1, with no augmentation from target domain) has high generalization (AUC) for low base error inputs while generalizing poorly for high base error inputs. 
Annotating with base-consistent-sampling is thus a waste of budget as the base and simple finetuned model are already good on the low base error inputs.
Moreover since LLM-annotations are not perfect, augmenting with base-consistent-sampling introduces noise into the model, when the model already has a high accuracy. 

On the other hand, high base error examples, which are targeted by base-inconsistent-sampling, do have substantial room for AUC improvement when considering the original finetuned model. This indicates that LLM annotations should focus only on base-inconsistent-sampling inputs, as such annotations may be the most \textit{informative}.  
\begin{table}[h]
\centering
\resizebox{0.5\textwidth}{!}{%
\begin{tabular}{|l|ll|}
\hline
\multirow{2}{*}{Pair} & \multicolumn{2}{c|}{Similarity} \\
 &  Base & Finetuned \\
\hline

What is a good diet plan for a commuter that wants to gain weight ? & \multirow{2}{*}{Low} &  \multirow{2}{*}{High}\\
What food should I eat to gain weight ? & & \\
\hline

How can you determine the structure for glucose ? & \multirow{2}{*}{High} &  \multirow{2}{*}{Low}\\
How can you determine the structure for oxygen ? & & \\
\hline





\end{tabular}
}
\caption{\label{tab:textual_semantic_qualitative}
Quora test examples having high \metric, i.e. finetuned predictions are different from base model predictions. Base model captures lexical similarity while finetuned captures target semantics.}
\end{table}


\subsection{\metric metric}

Based on the experiments above, we find that when annotated with LLMs, high base error, or \textit{base-inconsistent} examples are the most informative for training. But base-inconsistent-sampling, as described above, is not practical since it requires knowledge of the ground truth labels of inputs. Hence in this section, we develop an approximate metric for quantifying the degree of base-inconsistency of unlabeled inputs. 

We use a metric which measures deviation of the finetuned NLP model from the base model, and call it \metricnospace, since it depends on the base model in addition to the finetuned model. 
For a input $x_i$ we define it as 
\begin{equation}
\label{eq:metric_def}
    z_{i}(f,f_0) = \mathit{Dev}(f(x_i),f_0(x_i))
\end{equation}
where $f_0$ is the base model, $f$ is the finetuned model and Dev is a measure of deviation. We use simple squared error in our work.
The intuition is that during the finetuning process with the goal of minimizing error, a model is more likely to  deviate from the base model's score on an input if the base model has high error on that input. Here we assume that the finetuned model's score deviation captures this notion of base error, which can be generalized to the unlabelled inputs.

We present qualitative examples from our metric on Quora dataset in Table~\ref{tab:textual_semantic_qualitative}. These inputs were selected by our \metric metric as having high deviation. While for the first pair of examples the lexical similarity (base semantic) is of the pair is low, their semantic meaning (\textit{duplicate question} semantics) is the same, while for the second pair, while the lexical similarity is high, their semantic similarity is low. When doing LLM annotations, inputs like these would be the most informative for training. 

The formulation above defines \metric based on deviation of individual input semantic similarity scores. But we can also define \metric using deviation at a domain level. For example, for a multi-domain dataset with domain information for each input, the metric can be averaged over the entire domain to find the most suitable domains for LLM annotations. 

\section{EAGLE: Enhanced Generalization using LLM Annotations}
\label{sec:method}

Based on the \metric metric, we now present the \textit{EAGLE} algorithm for enhancing generalization of NLP models using LLM annotations. 
As in Section 3, we consider an active learning setup where we are given some labeled examples $L$ and a pool of unlabeled inputs $U$ along with a budget $B$ of annotating unlabeled inputs (using LLMs).  
In addition to standard classification tasks, our algorithm can also work for other tasks such as ranking. We first present the general algorithm and then present instantiations of it for a classification task (semantic similarity) and a ranking task (semantic search). 

\subsection{The EAGLE Algorithm}
\label{subsec:finalalgo}
\paragraph{Step 1: Computing \metric}
As the first step, we finetune our base model 
on the labeled data $L$ to get a finetuned model $f$ i.e.,
\begin{equation}
\label{eqn:finetuning}
    f = \mathrm{argmin}_f \mathbb{E}_{(x_i,t_i)\in L}[\mathcal{L}(f(x_i),t_{i})]
\end{equation}

Using $f$, we  compute the \metric score $z_{i}(f,f_0)$ where $f_0$ is the base model, for each unlabeled input $x_i \in U$ i.e.
\begin{equation}
z = \{z_i(f,f_0) : x_i \in {U} \}
\end{equation}

\paragraph{Step 2: Sampling inputs using \metric}
The next step involves sampling appropriate inputs for LLM annotations. We either choose to do an input-wise \metric sampling, or if the data is domain annotated, we can do domain level annotations. For input-wise sampling we select the top $B$ samples i.e. 
\begin{equation}
U_{sampled} = \{x_i : z_i \in \mathrm{top}(z,B)\}
\end{equation}

For domain level annotations, we can obtain the domain-level \metric metric (by averaging the metric over inputs belonging to the domain). 
In this case, the budget B is uniformly distributed over inputs in selected domains.


\paragraph{Step 3: Annotating sampled inputs using LLM}
Given a sampled set of unlabelled input $U_{sampled}$, we use LLM annotations for these inputs to get an annotated set as $L_{sampled}'$. 
We denote LLM annotations function by $\mathcal{T}' : \mathcal{X} \to \{0,1\}$, and hence the LLM annotation for input $x_i$ as $t'_{ij} \in \{0,1\}$. The augmented dataset made from $U$ is hence $L'$
\begin{equation}
   L_{sampled}' = \{(x_i,t'_{i}) : x_i \in U \} 
\end{equation}

\paragraph{Step 4: Finetuning classifier on augmented labelled data}
Finally we finetune the base model on the augmented dataset $L+L_{sampled}'$ using Eq~\ref{eqn:finetuning}.

\subsection{Application: Semantic Similarity}
\label{subsec:similarity_method}
We present how our algorithm can be used for the semantic similarity task described in Section 3.  
Step 1 follows from the main algorithm. The \metric computation follows Eq~\ref{eq:metric_def}, with the only caveat being that the classifier function now takes two inputs : 
\begin{equation}
    z_{i}(f,f_0) = \mathit{Dev} (f(x_i,y_i),f_0(x_i,y_i))
\end{equation}
Sampling is done in the same way with the algorithm selecting top $B$ inputs having highest $z_{i}$.

\paragraph{LLM Annotation Details}
Consider a set of unlabeled examples $U$ consisting of pairs $(x_i,y_i)$ to be annotated by the LLM. 
We construct a prompt which consists of set of pairs $(x_i,y_i)$ of sentences. For cost-efficiency we consider 10 pairs in each prompt for our experiments. The prompt asks the LLM to output all the pairs which are semantically similiar (with semantics defined appropriately inside the prompt). All pairs outputted by LLM as similar are considered similar while rest are not. 
See Table~\ref{tab:gpt_qualitative} for example annotation outputs on Quora dataset. 
\begin{table}[h]
\centering
\resizebox{0.5\textwidth}{!}{%
\begin{tabular}{|l|lll|}
\hline
Pair & Deviation & GT & LLM  \\
\hline

Why does Cuba tolerate the presence of Guantanamo Bay Naval Base ? & \multirow{2}{*}{Low} &  \multirow{2}{*}{0} &  \multirow{2}{*}{1}\\
What is the deal with Guantanamo Bay ? Why isn't it closed yet ? & & & \\
\hline

What are the best headhunters in Mexico ? & \multirow{2}{*}{Low} &  \multirow{2}{*}{1} &  \multirow{2}{*}{0}\\
Who are the best headhunters in Mexico ? & &  &\\
\hline

What is third pricing model ?& \multirow{2}{*}{High} &  \multirow{2}{*}{0} &  \multirow{2}{*}{0}\\
What is a pricing model ? & &  &\\
\hline

How was trading performed in Ancient India ? & \multirow{3}{*}{High} &  \multirow{3}{*}{1} &  \multirow{3}{*}{1}\\
Is there proof of ancient Indians trading overseas ? If yes , & &  &\\
then what did they trade in and with what countries ?  & &  & \\
\hline

\end{tabular}
}
\caption{\label{tab:gpt_qualitative}
LLM (gpt-3.5-turbo) annotations for some low and high deviation examples. LLM is able to correctly guess the high deviation ones while is incorrect on the low deviation ones. LLM annotation accuracy is agnostic of the deviation. See Supp~\ref{app_sec:llm_prompting} for prompts used.}
\end{table} 


\subsection{Application: Semantic Search} 
\label{subsec:search_method}
While semantic similarity is a fundamental task, real world applications often rely on \textit{semantic search}. 
In such applications, the $X$ is called set of all \textit{queries} denoted as $X = \{x_0,x_1,x_2,\ldots,x_{|X|}\}$, while $Y$ is the set of \textit{labels} denoted as $Y = \{y_0,y_1,y_2,\ldots,y_{|Y|}\}$.
These search for an optimal semantic match for a sentence $x \in X$ from the set $Y$, i.e. 
\begin{equation}
\label{eq:anns}
g(x,\mathcal{T}) = \mathit{argmax}_{y_i \in Y} \mathcal{T}(x,y_i)
\end{equation}
In practice since we don't have the true semantics $\mathcal{T}$ (e.g., relevance to query),  we use some approximation of semantics for argmax. 
We denote a set of examples by: 
\begin{equation}
    D_{\mathit{search}} = \{((x_i,Y),T_i) : i \in [|X|]\}
\end{equation}
where
\begin{equation}
\label{eq:t1}
    T_{i} = \{t_{ij} : j \in [|Y|] \}
\end{equation}
Unlabeled samples lack $T_i$ information. Following Eqn.~\ref{eq:metric_def}, \metric on set ${X}$ is defined as 
\begin{equation}
\begin{split}
\label{eq:metric_def_search}
y =&  g(x_i,f_0) \\
z_{i}(f,f_0) =& \mathit{Dev}( f(x_i,y), f_0(x_i,y))
\end{split}
\end{equation}
where $g(.)$ finds the nearest $y_j \in {Y}$ for $x_i$ according to base embedding function $f_0$ (Eqn~\ref{eq:anns}). 


\paragraph{LLM Annotation Details}
The unlabeled set $U$ now consisting of pairs $(x_i,Y)$ to be annotated by LLMs. 
Querying semantic similarity for each query, label pair  $\{(x_i,y_j) : y_j \in Y\}$ is very expensive. Hence we first create a filtered set of labels from a semantic similarity model (in our case finetuned model) $f$. With slight abuse of notation, we consider an extension of the function $g$ in Eq~\ref{eq:anns} as $g(x,f,K)$ where $g$ now outputs a set of top $K$ labels for each query. Our filtered set is hence $Y' = g(x,f,K)$ where $f$ is the finetuned model. 
The set $Y'$ hence consists of top $K$ ranked labels for a query according to the finetuned model $f$. The rest of the labels (which weren't in the top $K$ ranking of the finetuned model) i.e. 
$Y/Y'$ have their semantic similarity set to 0. We query the LLM for semantic similarity of labels in the filtered set $Y'$, where $|Y'| = K$. Hence this helps us reduce the complexity of searching through the whole label space by restricting the search space using the finetuned model $f$.
We take $K$ = 10 for all experiments. For each pair $\{(x_i,y_j) : y_j \in Y'\}$ we can then query LLM similar to semantic similarity setup above. 
\begin{equation}
\label{eq:t2}
   L' = \{((x_i,Y),T'_{i}) : (x_i,Y) \in U \} 
\end{equation}
We empirically observe that it is better to provide one prompt for each  query along with its top $K$ filtered labels. The labels should be ordered by their semantic similarity score according to the model $f$ in the prompt. Example prompts used in our experiments can be found in  Supp.~\ref{app_sec:llm_prompting}. 
The filtering step in semantic search can use any good similarity model. In our experiments, we utilise our finetuned model $f$ for the filtering step. 


\section{Experiments}
\label{sec:experiments}

\begin{figure*}[h]
\begin{subfigure}{0.48\textwidth}
\includegraphics[width=\textwidth]{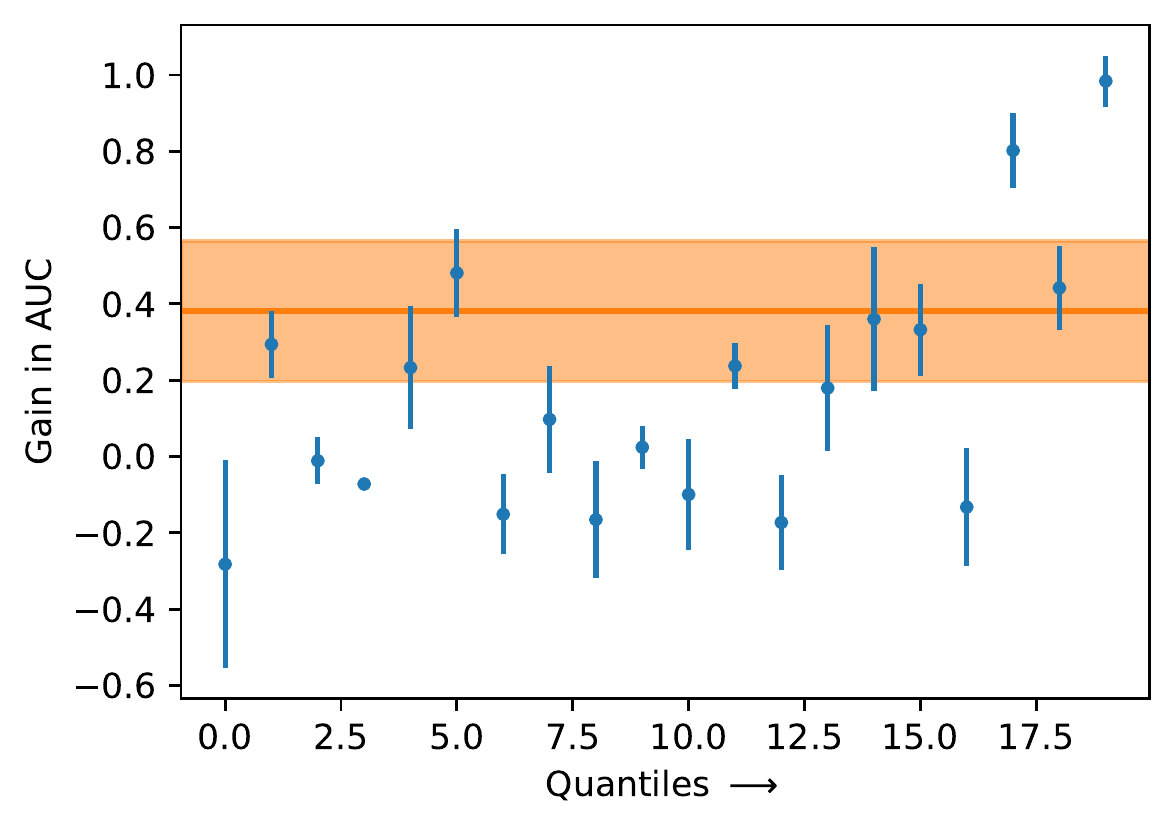}
\caption{LLM-annotated}
\label{fig:quantilewise_gpt_5percent}
\end{subfigure}
\begin{subfigure}{0.48\textwidth}
\includegraphics[width=\textwidth]{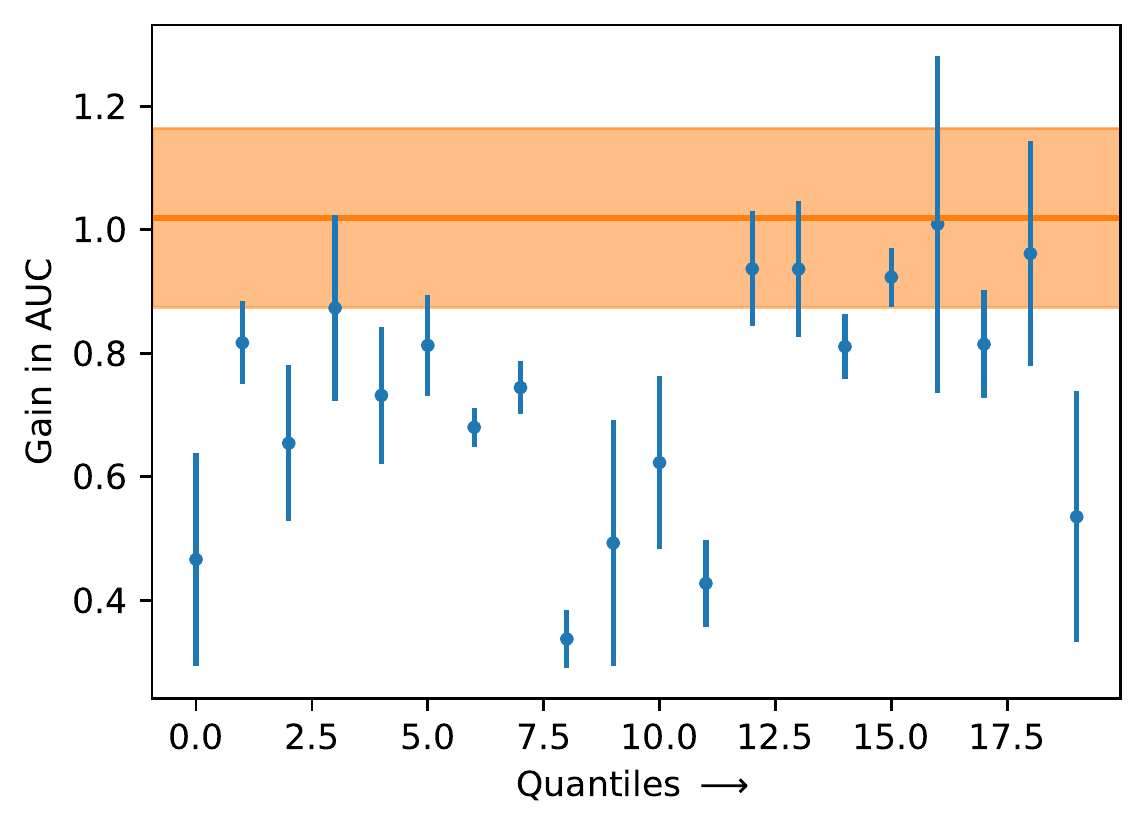}
\caption{GT-annotated}
\label{fig:quantilewise_gt_5percent}
\end{subfigure}
\caption{\label{fig:quantilewise_5percent} Gain in AUC on including LLM-annotated and GT-based augmentations on Quora dataset. The orange line is gain in AUC with uncertainty based sampling (shaded region shows std err.). We divide the unlabeled data into 20 quantiles, based on the \metric metric. \metric increases from left to right. For LLMs, uncertainty is not a good method for sampling and \metric based sampling is better, while for GT-based augmentations uncertainty based sampling is better.}
\end{figure*}


\begin{table}[t]
\centering
\resizebox{0.5\textwidth}{!}{%
\begin{tabular}{|l|r|}
\hline
Data & Test AUC   \\
\hline
Initial Train set  & 85.780 $\pm$ 0.002\\
\hline
+ Random LLM & 86.001 $\pm$ 0.139\\
+ Uncertainty LLM & 86.058 $\pm$ 0.089 \\
+ \metric LLM & \textbf{86.432 $\pm$ 0.106} \\
\hline
+ Random GT & 86.677 $\pm$ 0.068\\
+ Uncertainty GT & {87.125 $\pm$ 0.135} \\
+ \metric GT & \textbf{87.445 $\pm$ 0.091} \\
\hline
\end{tabular}
}
\caption{\label{tab:quora_hardvuncertain}
AUC for different sampling techniques for Quora semantic similarity task. We sample 10\% of unlabeled data. For LLM-based annotations, the best method is to sample using our \metric sampling while for GT based annotation, both uncertain and \metric based sampling are good.}
\end{table}

We evaluate EAGLE algorithm on two tasks: \textbf{1)} \textit{semantic similarity}, a fundamental task; \textbf{2)} \textit{semantic search}, a real-world task motivated by information retrieval applications. 
We assume that in addition to some labeled examples, we are also given a large pool of unlabeled inputs. For semantic similarity we consider generalisation in limited labeled data setting, while for semantic search we evaluate generalisation to unlabeled target domains. 
In limited labeled data setting, both the labeled and unlabeled inputs follow the same distribution while when adapting to unlabeled target domain (the unlabeled input), there is a distribution shift in inputs of the labeled and unlabeled examples. 
We show how our \metric based sampling inputs helps to improve generalization in both of these settings. 
We do our experiments on embedding-based semantic similarity as defined below.
\paragraph{Embedding-based Semantic Similarity/Search} 
The search argmax operation (Eq~\ref{eq:anns}) over the complete query set $|X|$ is quadratic (i.e. $|X|\times|Y|$).
For efficient computation, we use embeddings based semantic similarity (SBERT~\cite{reimers2019sentence}), where both queries and labels are seperately embedded into a $N$ dimensional unit norm space. The dot product  between the embedded representations of sentences gives the semantic similarity.
Hence the goal is to learn a embedding function $h : \mathcal{X}\cup\mathcal{Y} \to \mathbb{R}^N$ s.t. $h(x_i)^\intercal h(y_i)$ gives the semantic similarity between the functions. 

\paragraph{Sampling methods.} In addition to \metricnospace, we consider random-sampling and an active learning uncertainty-based sampling algorithm. As an oracle, we also consider ground-truth labels for the same inputs sampled by each of these sampling algorithms.

\paragraph{Implementation details.} We consider the base model as \texttt{MSMARCO-DistilBERT-v4} for both tasks. For LLM-based annotations we use GPT-3.5-Turbo. See Supp.~\ref{app_sec:llm_prompting} for prompts used in the experiments. We tried open source models such as \texttt{TogetherComputer/RedPajama-INCITE-7B-Base} (along with the Chat and Instruct version) or \texttt{MosaicML/mpt-7b-chat} but did not obtain good annotation accuracy.
All results are reported for 3 seeds. Other training details are in Supp.~\ref{app_sec:sstd}.

\begin{table*}[h]
\centering
\resizebox{\textwidth}{!}{%
\begin{tabular}{|l|ll|ll|}
\hline
 & \multicolumn{2}{c|}{Wikipedia}  & \multicolumn{2}{c|}{Amazon} \\
 & USA & Total & Books & Total \\
\hline
Initial Train set & 12.530 $\pm$ 0.034 & 19.048 $\pm$ 0.019 & 17.226 $\pm$ 0.008 & 24.904 $\pm$ 0.076 \\
\hline
+ Target LLM Random 40\%  & 13.188 $\pm$ 0.073 & 19.232 $\pm$ 0.024 & 17.959 $\pm$ 0.075 & 25.065 $\pm$ 0.049 \\
+ Target LLM \metric (bottom 40\%)  & 13.089 $\pm$ 0.079 & 19.209 $\pm$ 0.034 & 18.021 $\pm$ 0.033 & 25.110 $\pm$ 0.038 \\
+ Target LLM \metric (middle 40\%)  & 13.166 $\pm$ 0.021 & 19.228 $\pm$ 0.022 & 18.123 $\pm$ 0.060 & 25.216 $\pm$ 0.012 \\
+ Target LLM \metric (top 40\%) & \textbf{13.372 $\pm$ 0.058} & \textbf{19.363 $\pm$ 0.023} & \textbf{18.351 $\pm$ 0.028} & \textbf{25.271 $\pm$ 0.030} \\
\hline
+ Target GT Random 40\%  & 13.893 $\pm$ 0.047 & \textbf{19.430 $\pm$ 0.052} & 18.375 $\pm$ 0.068 & \textbf{25.329 $\pm$ 0.051} \\
+ Target GT \metric (bottom 40\%) & 13.911 $\pm$ 0.032 & 19.395 $\pm$ 0.013 & 18.455 $\pm$ 0.054 & 25.271 $\pm$ 0.020 \\
+ Target GT \metric (middle 40\%)  & \textbf{13.973 $\pm$ 0.070} & 19.327 $\pm$ 0.023 & 18.400 $\pm$ 0.027 & 25.213 $\pm$ 0.075 \\
+ Target GT \metric (top 40\%)  & 13.878 $\pm$ 0.015 & 19.414 $\pm$ 0.015 & \textbf{18.613 $\pm$ 0.043} & 25.285 $\pm$ 0.057 \\
\hline
\end{tabular}
}
\caption{\label{tab:wiki_amazon_ood} P@1 for test target domain (USA in Wikipedia and Books in Amazon) and complete test set. For LLM-based annotation, top 40\% samples according to our \metric are optimal for total accuracy (while also being optimal for target domains accuarcy). For GT based annotations, Random Sampling is best for total accuracy. Best target domain accuracy method for GT is inconclusive.}
\end{table*}

\subsection{Semantic Similarity}

\paragraph{Setup}
We conduct experiments on the \textit{Quora Question Pairs} ~\cite{wang-etal-2018-glue} dataset, which consists of pairs of questions. The task is to label each pair as a duplicate or not, i.e., whether the questions have the same intent or not. We subsample 38400 training pairs from the train set. 
We consider a setup where 10\% of the Quora dataset is labeled by ground truth, while rest of the 90\% forms the unlabeled pool of data. We present test AUC (Area-Under-ROC) numbers as the evaluation metric.

\paragraph{Comparison with Random and Uncertainty Sampling}
We follow the Algorithm from ~\ref{subsec:finalalgo} for semantic similarity. Using the model finetuned on labeled data, we sample 10\% of unlabeled data for annotation, according to various sampling strategies (namely random, uncertainty and \metricnospace). 
For details on how LLM annotations are done see Sec~\ref{subsec:similarity_method}.
We also present results on annotations with ground truth labels i.e. $t'_{i} = t_{i}$ (Sec~\ref{subsec:similarity_method}). 
In Table~\ref{tab:quora_hardvuncertain}, we show that for LLM-based annotations, \metricnospace -based sampling achieves significantly better test AUC than random and uncertainty sampling.
In comparison, for annotating with GT labels, both uncertainty and \metricnospace - based sampling yield high AUC.

\paragraph{Evaluating \metricnospace-based Quantiles} 
To find out why uncertainty-based sampling did not work for LLM annotations, we divide the data into 20 quantiles, each having 5\% of unlabeled data based on \metric metric. 
Figure~\ref{fig:quantilewise_5percent} shows the gain in AUC on including these samples (LLM or GT annotated) with the training data. As a comparison, the orange line in the plot signifies accuracy on sampling 5\% from uncertainty metric (shaded portion is std error). For LLM annotations, we observe that  uncertainty is not a good technique for sampling and \metricnospace-based sampling is better, while for GT-based augmentations uncertainty-based sampling provides better gains than  \metricnospace -based sampling.





\subsection{Semantic Search}

Next, we evaluate the utility of \metric sampling for generalisation to unlabeled target domains in semantic search tasks.
\paragraph{Datasets} We consider two recommendation datasets for semantic search : 1) \texttt{LF-WikiSeeAlsoTitles-320K}~\cite{Bhatia16} (i.e., Wikipedia) considers a recommendation/retrieval setting. The train set consists of Wikipedia page titles (queries $X$) along with a large set of page titles (labels $Y$).  
For Wikipedia, a label $y_j$ is semantically similar to a query $x_i$ if the label is likely to occur in the \textit{SeeAlso} section of the query article's wiki-page. As described in Section 4.3 for the semantic search task, the set of labels remains fixed to $Y$. The task is to learn embeddings which follow the semantics above.
For each article $X$ we also parse its category information, which we use as it domain label. If for an article $x_i$, it's categorical information contains "USA" or "America" it belongs to the domain \texttt{USA}, otherwise not.  
2) \texttt{LF-AmazonTitles-131K}~\cite{Bhatia16} (i.e.,  Amazon) considers recommendations in e-commerce \textit{AlsoBought} product setting. Given a query product ($X$) the labels correspond to possible products a user might buy ($Y$). 
Here too we consider categorical information for all query products $X$. We construct two domains in Amazon. All products in "Books" category are in the \texttt{Books} domain, while all products in the "Kitchen and Dining" category form the \texttt{Kitchen} domain. 

\paragraph{Setup} For Wikipedia we consider the \texttt{USA} domain as our target unlabeled domain, and the rest of the dataset as our labeled data. 
Similarly for Amazon we construct two versions of the dataset, one where we \texttt{Books} domain as the target unlabeled domain and another where we consider \texttt{Kitchen} as the target unlabeled domain. 
We use \textit{Precision@1} (P@1) metric for evaluation, i.e. the fraction of queries whose top ranked label is semantically similar (or \textit{relevant}) to the query, i.e, 
$$
\mathit{Precision@1} = \mathbb{E}_{x_i \in X}[\mathcal{T}(x_i,g(x_i,f))]
$$
For the GT annotation oracle, we annotate the top $K$ sampled labels (using the finetuned model $f$) with ground truth information i.e. for a query $x_i$, $t'_{ij} = t_{ij} \ \forall\  y_j \in Y'$ and $t'_{ij} = 0 \ \forall\ y_j \notin Y'$. See Sec~\ref{subsec:search_method} for notation (Eq~\ref{eq:t1},\ref{eq:t2}).
Note that for all labels which are not ranked in top $K$ by the finetuned model have their semantic similarity set to 0, even if they were relevant in GT. 
For other details refer to Supp.~\ref{app_sec:sstd}.


\paragraph{Results} We present test P@1 for the target domains (\textit{USA} domain in Wikipedia and \textit{Books} domain in Amazon) and the complete source + target domain  test sets in Table~\ref{tab:wiki_amazon_ood}. We find that when augmenting with LLM based annotations, selecting inputs which are in the top 40\% inputs according to our \metric are optimal for total accuracy (while also being optimal for target domains accuarcy). For GT based annotations, Random Sampling is best for total accuracy, though results are not significant. 

\paragraph{Using Domain-knowledge for Qualitative measure of \metric} 
On the Amazon recommendation task, consider domain adaptation to \texttt{Books} or \texttt{Kitchen} domains.  
For \texttt{Book} recommendations using only book titles (e.g., say \texttt{The Kite Runner} for  \texttt{A Thousand Splendid Suns}) the \metric would be high for encoder based models (assuming that encoder doesn't have the necessary domain knowledge for book recommendations, i.e., the two books share the same author). That is, it would require more world knowledge than for domains like \texttt{Kicthen}, (e.g., \texttt{Kaiser Bakeware Muffin Pan} for \texttt{Nordic Ware Brownie Pan}) which are more likely to be consistent with the base model's semantics (in this case lexical similarity).

For the domain Kitchen, we can see in Table~\ref{tab:distshift_kd} that including LLM-based annotations for domain Kitchen does not provide any gains compared to the base model. In comparison, for other domains like Books, LLM annotations lead to better generalisation than both base and training set finetuned models. Refer to Supp.~\ref{app_sec:amazonkitchen_llm_acc} for a plot showing how LLMs are not better than finetuned/base model for Amazon(Kitchen) domain, whereas for Wiki(USA) and Amazon(Books) LLMs are significantly better (Fig~\ref{fig:llm_acc}).
For accuracy improvements on Kitchen domain, techniques utilising regularisation to base model may be suitable and LLMs may not be needed.
\begin{table}[t]
\centering
\resizebox{0.5\textwidth}{!}{%
\begin{tabular}{|l|lll|}
\hline
Finetuning Dataset & Wikipeda(USA) & Amazon(Books) & Amazon(Kitchen) \\
\hline
\metric & 9.57 & 9.48 & 8.62\\
\hline
Base Model & 12.33 & 16.78 & \textbf{32.91}\\
Training set  & 12.54 & 17.37 & 31.53\\
Training set+Target LLM & \textbf{12.95} & \textbf{18.28} & \textbf{32.91}\\
\hline
\end{tabular}
} 
\caption{\label{tab:distshift_kd} Domain averaged \metric (x100) scores for the target domains  (USA in Wikipedia and Books/Kitchen in Amazon) are shown in the top row.
The next three rows present P@1 numbers for target domains. LLM annotations help for \textit{USA} and \textit{Books} target domain;  but for the \textit{Kitchen} target domain, the Base model has the same accuracy as LLM-augmented model.
}
\end{table}

\section{Conclusion}

We showed how LLMs can be used for annotations and how sampling of inputs plays an important role in improving an NLP model's generalization. To this end, we presented a novel sampling algorithm for input selection that performs better than the popular technique of uncertainty-based sampling.

As future work, we would like to test whether the \metric metric applies to other NLP tasks beyond semantic similarity. For the semantic search setting, given the generative capabilities of LLMs, an interesting future direction is to use LLMs to generate labels for queries  while restricting the generated label set to our target label set. 

\bibliography{anthology,custom}
\bibliographystyle{acl_natbib}

\appendix






\section{LLM Prompting}
\label{app_sec:llm_prompting}

\subsection{Semantic Similarity}
\paragraph{Prompts used : Systems and Question prompt}
See Table~\ref{tab:prompts_quora}
\begin{table*}[h]
\centering
\begin{tabularx}{\textwidth}{|l|X|}
\hline
System Prompt & You are an expert in judging the intended answer for short questions on quora. You use the intended answer to find duplicate questions (i.e. questions having the same intended answer and answering one of the questions will answer the other question too.
Given a pair of questions from one of these forums you are effectively able to discern if they are duplicates of each other or not by reasoning about the intended answer.\\
Question Prompt & Given pairs of questions from (say) Quora, output the pair of questions that are asking the same question (i.e. have the same intended answer). Reason about the intended answer to solve this.
Here is an example of output "Pair1".

Pair1: \%s
Pair2: \%s
Pair3: \%s
Pair4: \%s
Pair5: \%s
Pair6: \%s
Pair7: \%s
Pair8: \%s
Pair9: \%s
Pair10: \%s

Pairs having duplicates are : \\
\hline
\end{tabularx}
\caption{\label{tab:prompts_quora}
Prompts Used for Quora}
\end{table*}

\subsection{Semantic Search : Top K sampling with Finetuned for Recommendations}
\paragraph{Prompts used : Systems and Question prompt}

See Table~\ref{tab:prompts_wikiusa}, Table~\ref{tab:prompts_amazonbooks}, Table~\ref{tab:prompts_amazonbooks}.

\begin{table*}[h]
\centering
\begin{tabularx}{\textwidth}{|l|X|}
\hline
System Prompt & You are an expert on United States centric Wikipedia articles and article titles. You are able to infer the content and context of an article accurately from the title of the article alone. Also, given a reference article title, you are able to accurately discern which articles should be in the 'SeeAlso' section of the reference article's wikipedia page.\\
Question Prompt & You are given the main article title and titles for possible "SeeAlso" articles. You have to output which <SeeAlsoArticle> is most likely to be in the "SeeAlso" section of the main article.
The topics of the articles are around or relating to United States somehow. 
You should infer the content of the articles from their titles and output SeeAlso articles are closely related to the main article's topic and provide additional useful information to the reader. 
SeeAlso articles should also help readers explore related areas and should have value to the reader. 
Output just the article id (e.g. SeeAlsoArticle1) for the most likely article. Here is an example of output "SeeAlsoArticle1". 

MainArticleTitle: \%s
SeeAlsoArticle1: \%s
SeeAlsoArticle2: \%s
SeeAlsoArticle3: \%s
SeeAlsoArticle4: \%s
SeeAlsoArticle5: \%s
SeeAlsoArticle6: \%s
SeeAlsoArticle7: \%s
SeeAlsoArticle8: \%s
SeeAlsoArticle9: \%s
SeeAlsoArticle10: \%s

Most likely <SeeAlsoArticle> to be in "SeeAlso" section is : \\
\hline
\end{tabularx}
\caption{\label{tab:prompts_wikiusa}
Prompts Used for Wikipedia USA}
\end{table*}

\begin{table*}[h]
\centering
\begin{tabularx}{\textwidth}{|l|X|}
\hline
System Prompt & You are an expert on relevance (/similarity) between books sold on e-commerce websites. Given a reference book, you are able to accurately discern the relevance of other books to the reference book.\\
Question Prompt & Given a product a customer has recently bought and a list of 10 possible products, output (product id of) which product is most relevant for the customer. 
Relevant products have similar/complementary use cases. Here is an example of output "Product1".

Question
BoughtProduct: \%s
Product1: \%s
Product2: \%s
Product3: \%s
Product4: \%s
Product5: \%s
Product6: \%s
Product7: \%s
Product8: \%s
Product9: \%s
Product10: \%s

Most relevant product is : \\
\hline
\end{tabularx}
\caption{\label{tab:prompts_amazonbooks}
Prompts Used for Amazon Books}
\end{table*}

\begin{table*}[h]
\centering
\begin{tabularx}{\textwidth}{|l|X|}
\hline
System Prompt & You are an expert on 'Kitchen and Dining' products sold on e-commerce websites and on judging their utility to a customer. Given a product bought from e-commerce website, you are able to accurately discern the relevance of other products to the customer.\\
Question Prompt & Given a product a customer has recently bought and a list of 10 possible products, output (product id of) which product is most relevant for the customer. 
Relevant products have similar/complementary use cases. Here is an example of output "Product1".

Question
BoughtProduct: \%s
Product1: \%s
Product2: \%s
Product3: \%s
Product4: \%s
Product5: \%s
Product6: \%s
Product7: \%s
Product8: \%s
Product9: \%s
Product10: \%s

Most relevant product is : \\
\hline
\end{tabularx}
\caption{\label{tab:prompts_amazonkitchen}
Prompts Used for Amazon Kitchen}
\end{table*}

\section{LLM Annotation Accuracy}
\label{app_sec:amazonkitchen_llm_acc}

\begin{figure*}[h]
\centering
\begin{subfigure}{0.48\textwidth}
\includegraphics[width=\textwidth]{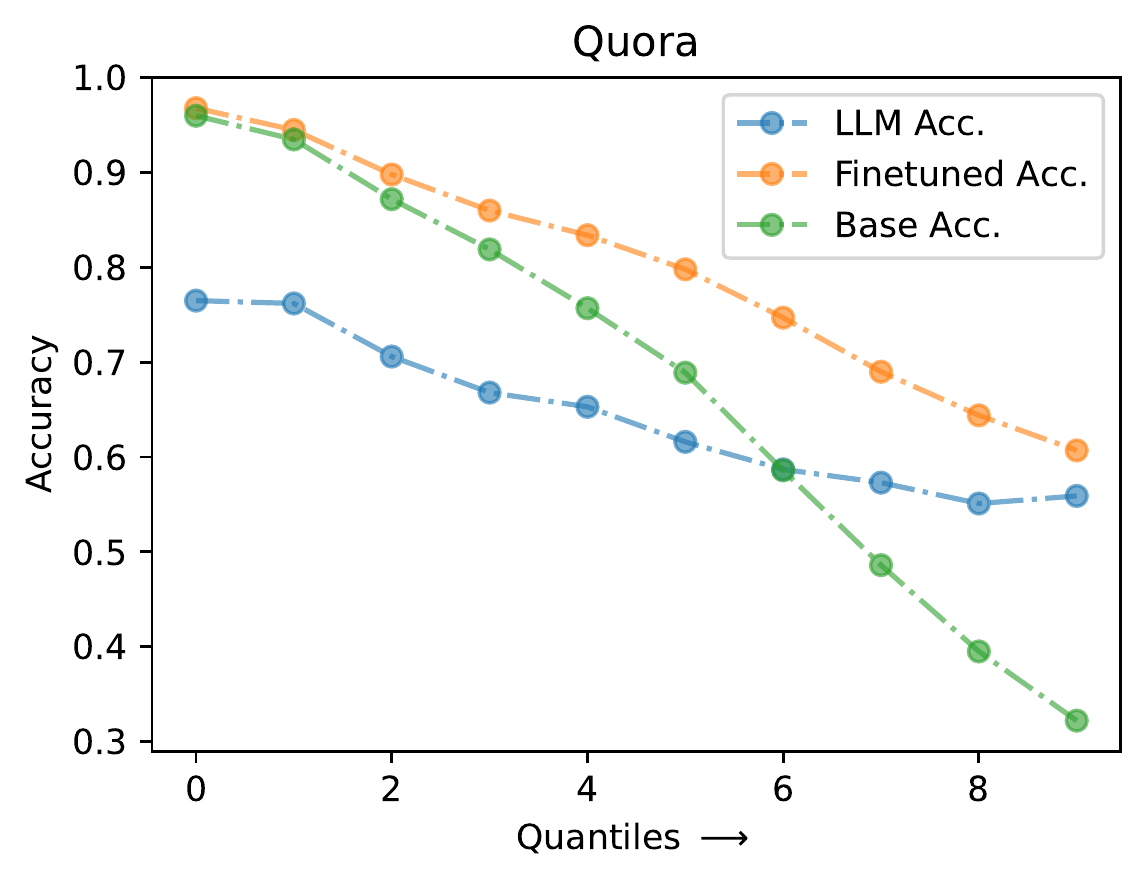}
\caption{LLM vs Finetuned accuracy across quantiles on Quora Question Pairs}
\label{fig:quora_llm_acc}
\end{subfigure}
\hfill
\begin{subfigure}{0.48\textwidth}
\includegraphics[width=\textwidth]{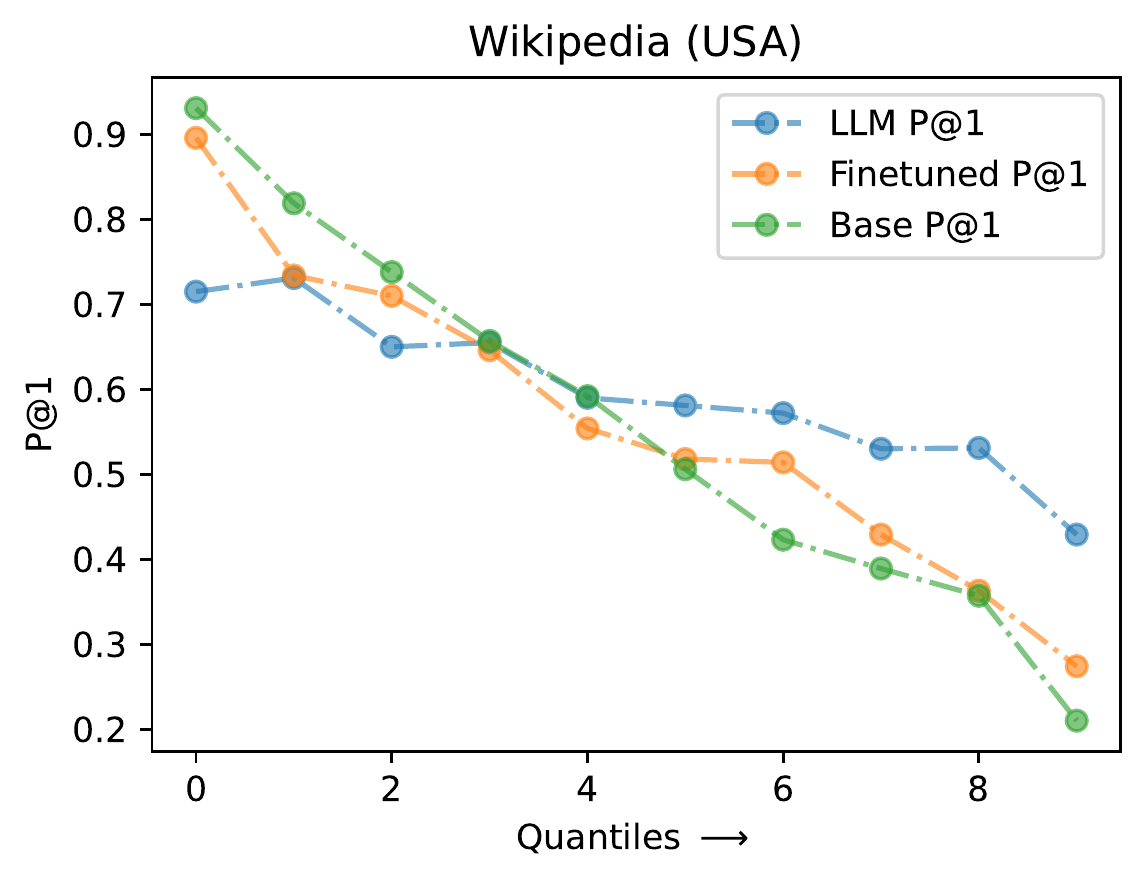}
\caption{LLM vs Finetuned accuracy across quantiles on Wikipedia USA category}
\label{fig:wikiusa_llm_acc}
\end{subfigure}
\vfill
\begin{subfigure}{0.48\textwidth}
\includegraphics[width=\textwidth]{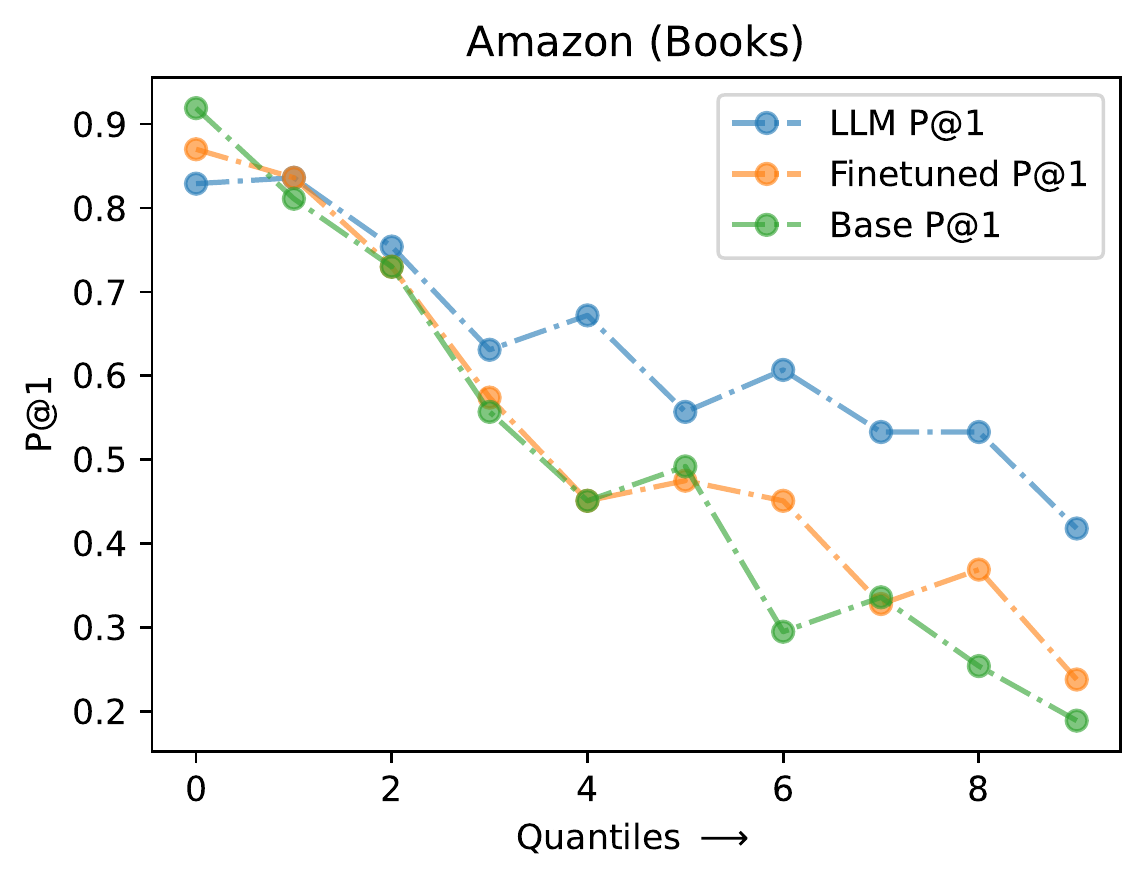}
\caption{LLM vs Finetuned accuracy across quantiles on Amazon Books category}
\label{fig:amazonbooks_llm_acc}
\end{subfigure}
\hfill
\begin{subfigure}{0.48\textwidth}
\includegraphics[width=\textwidth]{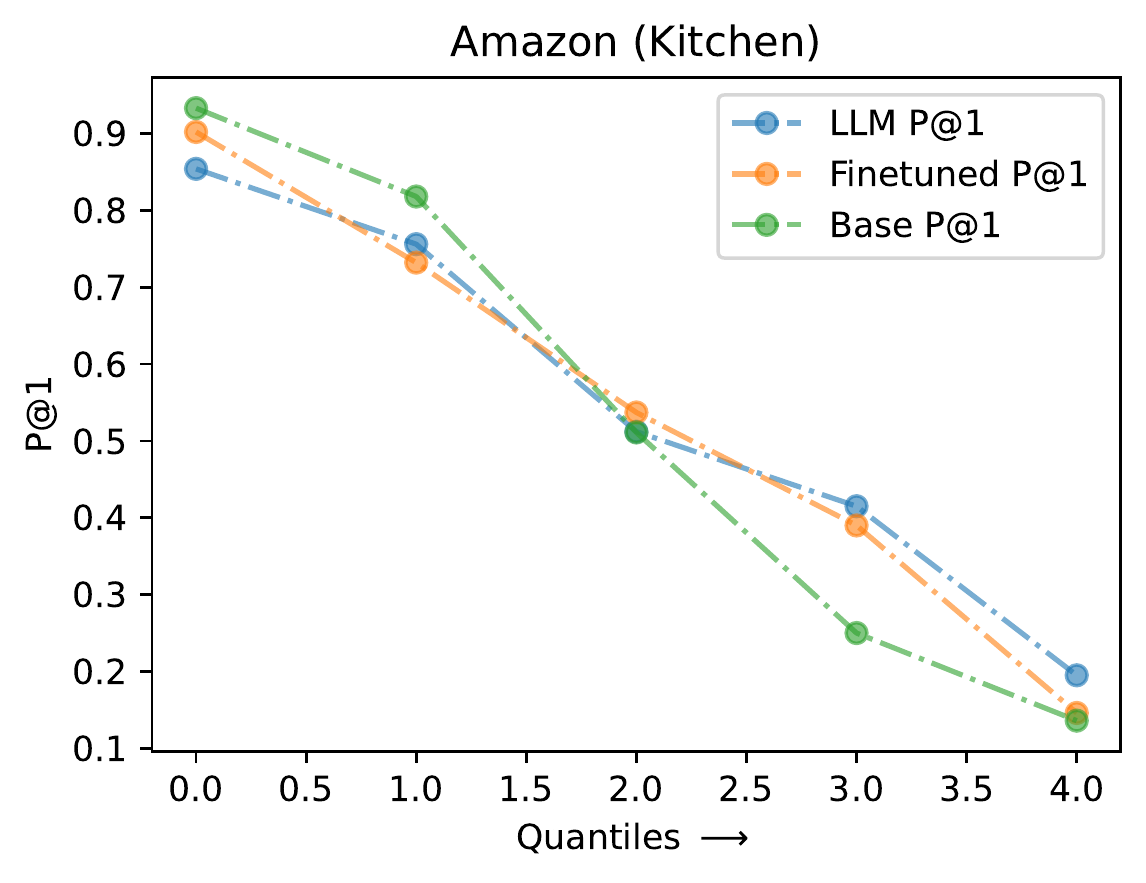}
\caption{LLM vs Finetuned accuracy across quantiles on Amazon Kitchen category}
\label{fig:amazonkitchen_llm_acc}
\end{subfigure}
\caption{LLM vs Finetuned accuracy across quantiles}
\label{fig:llm_acc}
\end{figure*}

For different \textit{Ground Truth \metricnospace-based} quantiles (i.e. the metric is computed over base model error with the ground truth label instead of a deviation with finetuned model) we show the accuracy of LLM annotations as compared to the Finetuned and Base models across different datasets. For Quora (Fig~\ref{fig:quora_llm_acc}) we can see that LLMs augmentation performance is always worse than finetuned model and is only comparable on the highest quantiles. Similarly for Wikipedia-USA (Fig~\ref{fig:wikiusa_llm_acc}) we can see that LLMs are better than finetuned model for around 50\% of higher quantiles while for Amazon-Books (Fig~\ref{fig:amazonbooks_llm_acc}) LLMs are always better than/comparable to finetuned model. For Amazon-Kitchen (Fig~\ref{fig:amazonkitchen_llm_acc}), LLMs are comparable to finetuned and base model and hence don't offer much advantage.
This analysis can be used by practitioner to decide which domains to augment with LLMs. 

\section{Training Details}
\label{app_sec:sstd}
\subsection{Semantic Similarity}
We finetune using quora with learning rate of 1e-4 for 2 epochs for all experiments. We use a batch size of 32 and have a linearly decay learning rate scheduler. We use MSE Loss for training. The initial quora dataset is subsampled by a factor of 10 (i.e. 38400 samples). We further subsample by 10 for active learning setup. 
\subsection{Semantic Search}
Recommendation are trained using a recent state-of-art algorithm ~\cite{dahiya2023ngame}.
We also subsample both the datasets by a factor of 10. We finetune both models for 100 epochs. 

\section{Ground Truth Annotations for Hard/Easy Target Domain in Quora}
\label{app_sec:quora_motivation_gt}

See Table ~\ref{tab:quora_motivation_gt}

\begin{table*}[h]
\centering
\begin{tabular}{|l|rrr|}
\hline
Data & base-inconsistent-sample & base-consistent-sample & Total  \\
\hline
Training set & 59.335 $\pm$ 0.139 & 99.068 $\pm$ 0.048 & 86.824 $\pm$ 0.038\\
\hline
+ 100\% (complete target domain) & 78.134 $\pm$ 0.187 & 98.229 $\pm$ 0.084 & 90.740 $\pm$ 0.063 \\
\hline
+ Random 50\%  & 74.432 $\pm$ 0.204 & 98.496 $\pm$ 0.063 & 89.893 $\pm$ 0.123\\
+ Uncertain 50\%  & 77.083 $\pm$ 0.248 & 98.077 $\pm$ 0.090 & 90.159 $\pm$ 0.069\\
+ base-consistent-sample 50\%  & 51.403 $\pm$ 0.390 & 99.499 $\pm$ 0.018 & 86.034 $\pm$ 0.046\\
+ base-inconsistent-sample 50\% & 79.801 $\pm$ 0.142 & 96.830 $\pm$ 0.067 & 89.644 $\pm$ 0.065\\
\hline
+ Random 16\% & 69.761 $\pm$ 0.352 & 98.732 $\pm$ 0.014 & 89.020 $\pm$ 0.071\\
+ Uncertain 16\% & 70.102 $\pm$ 0.164 & 98.522 $\pm$ 0.047 & 88.735 $\pm$ 0.060\\
+ base-consistent-sample 16\% & 56.082 $\pm$ 0.350 & 99.266 $\pm$ 0.040 & 86.561 $\pm$ 0.103\\
+ base-inconsistent-sample 16\% & 76.093 $\pm$ 0.325 & 97.709 $\pm$ 0.065 & 89.482 $\pm$ 0.098\\
\hline
\end{tabular}
\caption{\label{tab:quora_motivation_gt}
AUC for Quora with different Ground Truth (GT) Annotations.}
\end{table*}

\section{Formalising Semantic Similarity}
\label{app_sec:formanlising_ss}
Given a pair domains of sentences (short texts), $\mathcal{X}$ and $\mathcal{Y}$, the task of \textit{semantic similarity} is concerned with learning a function $f : \mathcal{X} \times \mathcal{Y} \to \{0,1\}$ which for a given pair of sentence $(x_i,y_j)$ from the set $\mathcal{X} \times \mathcal{Y}$ outputs either 1 (or 0) to show that $x_i$ is (or isn't) \textit{semantically} same as $y_j$. 
For symmetric tasks (like duplicate question detection) the set $\mathcal{X}$ and $\mathcal{Y}$ can be the same, but we consider the general case. 
The \textit{semantics} are broadly defined and depend on the target task. For e.g., for task of duplicate question detection (say on quora) a good model $f$ outputs whether for a pair of question the answer for one of the questions, answers the second question. For the task of say books recommendations, semantics might require capturing the similarity in pairs of sentences in context of the authors of the books, the books genre, their target audience etc. 
We hence mathematically denote semantics as $\mathcal{S} : \mathcal{X} \times \mathcal{Y} \to \{0,1\}$.

We consider the subset of these domains as $X = \{x_0,x_1,\ldots x_{|X|}\} \subseteq \mathcal{X}$, $Y = \{y_0,y_1,\ldots y_{|Y|}\} \subseteq \mathcal{Y}$.  The similarity between a pair $(x_i,y_j)$ is hence $\mathcal{S}(x_i,y_j)$ shortened as $s_{ij}$. 
For semantic search a set of examples can be sufficient represented by subset of pair of indices $D_0 \subseteq [|{X}|] \times [|{Y}|]$ denoted by 
\begin{equation}
    D = \{(x_i,y_j,s_{ij}) : (i,j) \in D_0 \}
\end{equation}
where $x_i,y_j$ are the pair of sentences and $s_{ij}$ is the semantic similarity. Unlabeled data lacks $s_{ij}$ information.

\end{document}